# Artificial Intelligence for Drug Discovery: Are We There Yet?

Authors: Catrin Hasselgren[1] and Tudor I. Oprea[2,3]

1. Safety Assessment, Genentech, Inc., South San Francisco, CA 94080, USA

2. Expert Systems Inc., San Diego, CA 92130, USA

3. Internal Medicine, UNM Health Sciences Center, Albuquerque, NM 87131

## Abstract

Drug discovery is adapting to novel technologies such as data science, informatics, and artificial intelligence (AI) to accelerate effective treatment development while reducing costs and animal experiments. AI is transforming drug discovery, as indicated by increasing interest from investors, industrial and academic scientists, and legislators. Successful drug discovery requires optimizing properties related to pharmacodynamics, pharmacokinetics, and clinical outcomes. This review discusses the use of AI in the three pillars of drug discovery: diseases, targets, and therapeutic modalities, with a focus on small molecule drugs. AI technologies, such as generative chemistry, machine learning, and multi-property optimization, have enabled several compounds to enter clinical trials. The scientific community must carefully vet known information to address the reproducibility crisis. The full potential of AI in drug discovery can only be realized with sufficient ground truth and appropriate human intervention at later pipeline stages.

## Table of Contents

## 1. Introduction

Drug discovery is a systematic scientific process that aims to identify, design, and develop novel therapeutic agents to cure, ameliorate or prevent diseases and medical conditions. Drug discovery is often called a 'pipeline,' which suggests a unidirectional transition from hit/lead to candidate and marketed drug, supported by basic and clinical research (1). The process is, in fact, iterative in nature, multi-faceted, and complex; for example, small-molecule drug discovery requires a) basic science research and target identification; b) target pharmacology and biomarker development; c) lead identification; d) lead optimization, candidate selection, IND (Investigational New Drug)-enabling studies and scale-up for manufacturing; e) clinical research and development; f) regulatory review; g) post-marketing and h) medical practice deployment (2). Two drug discovery, development, and deployment maps (4DM) that depict key steps in the drug development lifecycle, one for small molecules and one for biologics, illustrate the interdependencies and complexities of this process and are available for download at the National Center for Advancing Translational Sciences (NCATS) website (https://ncats.nih.gov/translation/maps). Drug discovery scientists use various techniques and methodologies, such as computational modeling, medicinal chemistry, high-throughput screening (HTS), and biological assays, to identify promising compounds and evaluate their safety, efficacy, and pharmacokinetics.

As therapeutic modalities continue to evolve (3), the drug discovery process is adapting by incorporating novel data science, informatics, and artificial intelligence (AI) methodologies, among other technologies, to improve efficiency, and reduce costs and animal experiments, thus accelerating the development of novel, effective treatments. The impact of big data and AI in drug discovery, the subject of a 2020 review in this journal (4), continues to attract significant interest: Investors (5, 6), industrial (7, 8) and academic scientists (9, 10), and legislators (https://bit.ly/40LRAip) discuss the impact of AI for drug discovery (AI4DD). As of July 3, 2023, "Artificial intelligence in drug discovery" is in the title of 63 publications since 2019, according to Google Scholar (https://bit.ly/3FWk3dH).

Successful drug approval requires concurrently optimizing multiple properties related to pharmacodynamics, pharmacokinetics, and clinical outcomes. Pharmacodynamic properties are related to drug-target interactions and efficacy; appropriate pharmacokinetics include absorption, distribution, metabolism, excretion, toxicity (ADMET), and drug safety; clinical outcomes include the therapeutic intent, as detailed on the list of drug indications and off-label uses, as well as undesired outcomes such as side effects or adverse drug reactions. Thus, a successful drug discovery launch relies on three pillars: Diseases, targets, and therapeutic modalities. AI impacts most therapeutic modalities, including targeted protein degradation (11), antibodies (12), gene therapy (13), and oligonucleotide (14) and vaccine (15) design. In this review, we explore the impact of AI4DD on these three pillars, primarily focused on small molecule drugs as therapeutic modality. Its impact on regulatory science and some caveats of AI4DD are also included.

## 2. Emergence of artificial intelligence for drug discovery

**The knowledge deficit.** One of the main challenges human investigators and AI systems face in drug discovery is harnessing large volumes of heterogeneous data of varying quality. The rapid expansion of data and computing power has been described as justifying "a fourth paradigm," a.k.a, "data-intensive scientific discovery (16). Specifically for the "why" and "what if" types of questions, relevant, preferably reliable data must be identified, inferred when absent, and connected through evidence-based reasoning, illustrated by the metaphor "connecting the dots" (17). It is increasingly clear that modern drug discovery requires computer-based systems that can intelligently reason and recognize patterns, i.e., AI. These AI systems must be able to weigh data elements and aggregate instances of those patterns to assess confidence and justification. Automated systems that digest large sets of data using named entity recognition, NER (18), are an integral part of public domain databases, for example, DISEASES for gene-disease associations (19); STRING for protein-protein interactions (20); and OpenTargets (21) and Pharos (22) for complex disease-protein-drug annotations, to name a few. Combined with AI-based systems that predict protein structures, such as AlphaFold (23) and RoseTTafold (24), these resources are poised to accelerate AI4DD.

**An incomplete history of AI for small molecule drug design:** The general notion that, at some point, an AI platform will drive drug discovery has probably been formulated in the early days of science fiction. For example, TV shows like *Star Trek* have featured medicine-specialized holograms (AI entities) solving medical emergencies. As early as 1981, the idea of "designing drugs with computers" was motivating scientists at Washington University in St. Louis (25) and Merck Research Laboratories (26). The *Journal of Computer-Aided Drug Design* was founded in 1987, and in almost every decade since the 1980s, the notion of "AI in drug discovery" has captured our imagination.

Machine learning (ML) models that discriminate "drugs" from "non-drugs" emerged in 1998, when scientists from Vertex (27) and BASF (28) independently proposed models to estimate "drug-likeness," showing that it is possible to train ML models based on chemical features to discriminate "drugs" (specifically, compounds that medicinal chemists have proposed for biological testing) from "non-drugs" (compounds lacking pharmaceutical use). The challenge faced by ML models of drug-likeness is illustrative of the general challenge in drug discovery (specifically, small molecule drugs): The "drug" quality is not an intrinsic property of chemicals since regulatory agencies and, implicitly, humans approve compounds for medicinal use (29). Furthermore, drug approval is subject to revision and can be withdrawn over time. While most market withdrawals are caused by drug toxicity (30), lack of efficacy or economic reasons can also result in withdrawal. Despite these caveats, ML-based drug-likeness quantification is an integral part of the drug discovery process, with nearly 20 independent papers dedicated to drug-likeness discussed elsewhere (29).

**Current impact in drug discovery.** An in-depth scientometric analysis (31) of AI4DD showed a steep rise in publications, from 49 in 2011 to 333 in 2020. The number of AI-based drug discovery platforms is anticipated to increase in the near future. There is frequently a synergistic relationship between the pharmaceutical and biotech industries, which propel AI-driven drug discovery towards commercial use, and academic institutions, which often spearhead the development of algorithms and methods. Over the past two decades, AI and machine learning have shifted from being peripheral technologies to occupying a central role in the drug discovery process. Today, we are closer than ever to achieving this long-sought goal.

## 3. AI applications in various stages of drug discovery

### 3.1. Diseases and therapy selection

**Diseases and healthcare:** The last decade has witnessed significant computational advances in disease diagnosis, with ten or more publications describing the use of AI in dermatology, tuberculosis, Alzheimer's disease, diabetes, hypertension, and cancer, reviewed elsewhere (32). The potential influence of AI in healthcare professions, such as radiology (33), pathology (34), clinical pharmacology (35) and COVID drug repurposing (36), has been a topic of discussion. Recently, ChatGPT, a large language model (LLM) from OpenAI (https://chat.openai.com/), achieved a notable milestone by scoring 60% proficiency on the United States Medical Licensing Exam, USMLE (37). AI has the potential to impact healthcare management in areas like diagnostics, imaging analytics, patient-provider interactions, hospital and nursing home management, population wellness using social determinants, and real-world evidence collection and analysis (38). As of 30 March 2023, the general opinion is that AI will support healthcare professionals but is not expected to replace them in the near future.

**Image-based AI**. AI systems based on image analysis have been developed in various medical fields, including radiology, pathology, and dermatology. Pathology is evolving into digital pathology, which involves digitizing histopathology, immunohistochemistry, and cytology slides and training AI systems on this digital data (39). Computational pathology aims to reduce diagnostic and classification errors while accelerating biomarker discovery, pathophysiology evaluations, animal research, and toxicological assessments (40). In radiology, AI can assist radiologists in improving diagnoses and enhancing patient care by analyzing chest X-rays, magnetic resonance images, and computed tomography scans (41). However, a systematic framework may be needed for comparing human and AI perception in medical diagnosis (42). An ML model based on convolutional neural networks (CNN) has demonstrated near-physician accuracy in diagnosing atopic dermatitis. Furthermore, this CNN model performs well in differentiating atopic dermatitis from other skin conditions such as impetigo, mycosis fungoides, herpes simplex, and Kaposi varicelliform eruption (43).

**Nosology.** Nosology, the ability to classify, recognize, cross-reference, and reconcile disease terms, poses significant challenge for healthcare AI systems. While numerous disease and phenotype ontologies exist (44), the Monarch Disease Ontology (Mondo) is emerging as a comprehensive, community-driven, and computationally-driven resource (45). A systematic analysis of resources integrated into Mondo revealed nearly 10,500 unique rare diseases, in contrast to the commonly cited 7,000 figure (46). The successful combination of human and machine learning (Bayesian) curation employed by Mondo serves as a clear example of how human and computer intelligence can collaborate in understanding diseases.

**Drug indications and therapeutic intent:** A critical aspect of drug discovery involves understanding the relationship between drugs and diseases, as specified on the drug label or drug indication, typically listed in the "indications and usage" section. Uses not stated on the package inserts are called "off-label" uses. As of January 2023, the DrugCentral database contains 2331 FDA-approved human drugs associated with 2644 drug indications and 866 off-label uses (47). Therapeutic intent, the rationale behind choosing a therapy and the context in which it is prescribed, is vital to medical practice. Although DrugCentral maps drug uses to existing terminologies like SNOMED-CT and MeSH, two global standards for health terms, capturing therapeutic intent computationally is challenging (48). Therapeutic intent encompasses disease concepts and contextual factors such as pre-existing conditions, co-prescribed medications, specific genotypes, and various phenotypic, anatomical, or temporal parameters. The state-of-the-art in drug indication curation from text was expanded through InContext (49), which includes therapeutic context information about drug usage, especially for antineoplastic and cardiovascular drugs. However, appropriately capturing and validating "off-label medical uses" still requires manual curation (50). Developing AI systems capable of processing, corroborating, and cross-mapping disease terminologies and therapeutic intent is a significant goal for automated drug discovery and repurposing *in silico*.

## 3.2. Target identification and validation

**Knowledge Graphs.** A crucial early step in modern drug discovery is target identification. This process involves discovering and validating the clinical relevance of a biomolecule that serves as mode-of-action for the therapeutic modality (51). Although novel drug targets are typically identified through genomic, proteomic or phenotypic experiments, machine learning is increasingly being explored as target identification technology (52), often by means of knowledge graphs (53). A knowledge graph (KG) is a knowledge base that uses graph-structured data models or topology to integrate data. Drug discovery KGs store interlinked descriptions of nodes – genes, phenotypes, or compounds – while also encoding the underlying relationships (edges). KG data projection enables network-based analytical algorithms, thus turning complex drug discovery data into ML-ready files. Biological system networks are heterogeneous with multiple node and edge types (Fig 1). Developments in heterogeneous (54) relationship predictions (55) introduced and formalized a new framework that takes into account network heterogeneity by defining type-specific node-edge paths or meta-paths (56).

A meta-path encodes type-specific network topology between the source node (e.g., Protein target) and the destination node (e.g., Disease or Function). Type-specific meta-paths are: (Target — (member of) → Protein-Protein Interaction (PPI) Network ← (member of) — Protein — (associated with) → Disease) and (Target — (expressed in) → Tissue ← (localized in) — Disease). Type-specific meta-path counts can be combined using, e.g., degree-weighted paths (55) to dampen the effect of highly connected nodes. By assembling heterogeneous data types such as those included in Pharos (22), an ML-ready KG can capture data from major areas specific for human protein-coding genes: phenotype and disease, pathways, and interactions. Each area has appropriate levels of data, e.g., expression, association, membership, treatments, localization, and gene signatures. Each instance of a meta-path represents a specific chain of evidence of associations between a source and a destination node.

**Novel targets for Alzheimer's disease.** The meta-path framework can be applied in conjunction with various classification algorithms. In this case, XGBoost, a form of extreme gradient boosting (67), was utilized to identify novel genes associated with Alzheimer's disease (68). By integrating the comprehensive Pharos dataset, a web-based interface for data collected by the Illuminating the Druggable Genome initiative, into XGBoost, an AD-focused model was developed, which included 53 positive (AD-associated) and 3,952 negative (non-AD-associated) genes. The model's validity was assessed using the top 20 and bottom 10 genes across three biologically distinct AD model systems (68). This approach uncovered five previously unassociated gene targets related to immunity in AD: *FRRS1, CTRAM, SCGB3A1, CIBAR2,* and *TMEFF2.* These genes demonstrated significant differences between AD and control samples. The model identified key variables associated with inflammatory processes and oxidative stress. Although constrained by input data that lacked non-human information, the AD-specific XGBoost model emphasized the role of infection as a critical factor in AD pathology. There is a growing body of evidence supporting the hypothesis that brain infection may be the primary cause of AD (69).

**Novel genes for autophagy.** Autophagy (ATG) is an intracellular physiological degradation process responsible for eliminating protein aggregates, pathogens, and other cargo by delivering them to lysosomes for degradation (70). Although the number of known genes involved in ATG is increasing (71), a systematic experimental method for identifying ATG-related genes has yet to be established (72). Given the complexity of ATG-associated genomic datasets, an ATG-specific XGBoost machine learning model was developed to aid in designing relevant experiments (73). The ATG-specific meta-path XGBoost model used 103 ATG-associated (positive) genes and 3468 negatives (not ATG-associated genes), collating data from Pharos. Among the top 251 genes predicted to have ATG associations, 43 genes were already in the Autophagy Database, and another 15 had potential ATG associations. However, 193 genes (77%) did not have any apparent connection to ATG. These 193 genes could be considered "ATG dark genes." A literature review of newly published experimental reports was used to validate 7 of the top 20 and 2 of the bottom predicted ATG dark genes (73).

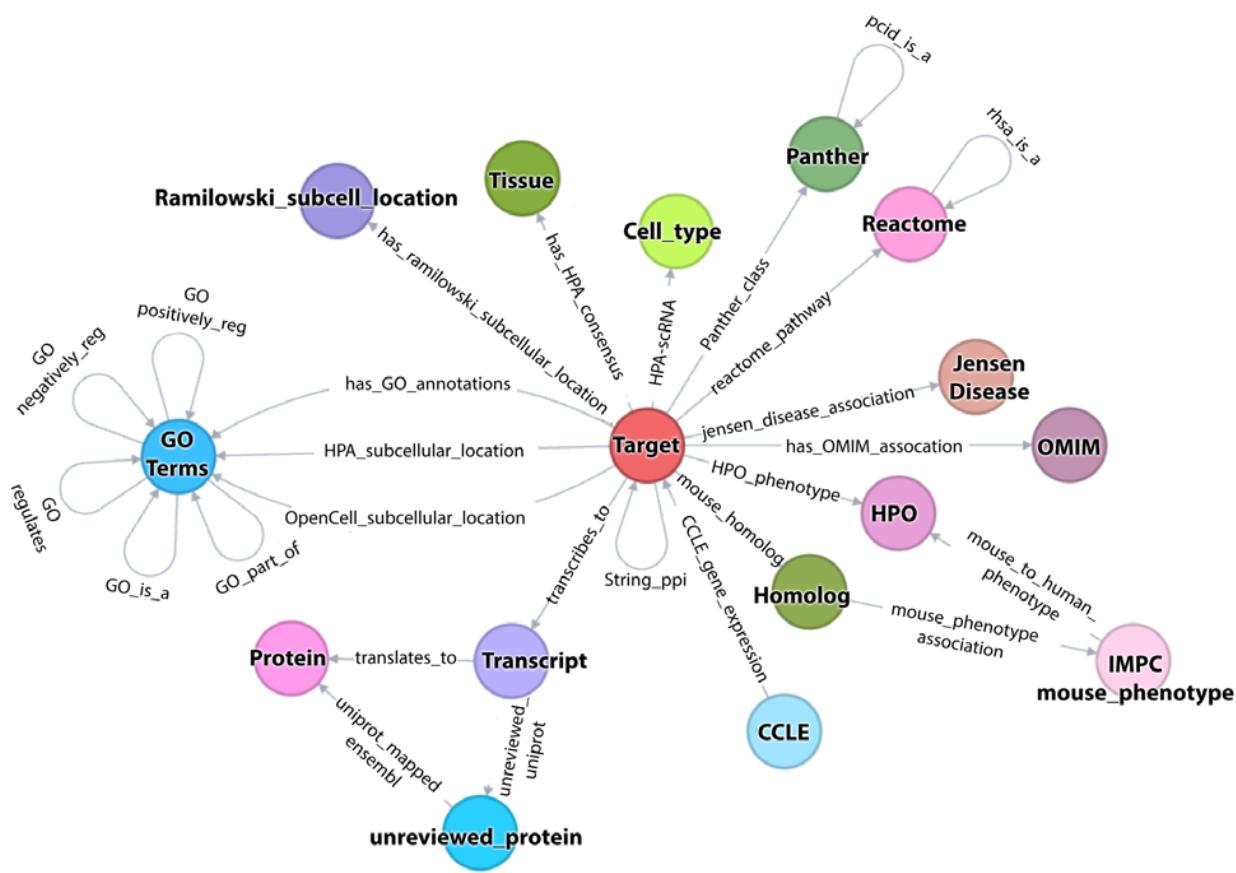

**Figure 1.** Knowledge Graph representation of phenotype-protein associations. Nodes (main entities) are related by edges (arrows). Centered on the main node, "Target", nodes listed clockwise are as follows: Tissue; Cell_type; Panther, Protein Analysis Through Evolutionary Relationships, a protein classification scheme(57); Reactome, a pathways database (58); Jensen DISEASES (19); OMIM, Online Mendelian Inheritance in Man database(59); HPO, human phenotype ontology (60); Homolog (which lists proteins with common evolutionary origin), further linked to IMPC (International Mouse Phenotype Consortium) mouse phenotype database (61); CCLE, cancer cell line encyclopedia (62); Transcript, Protein and unreviewed_protein – various types of UniProt (63) entries; GO, Gene Ontology (64) consortium terms; Ramilowski_subcellular_location, a specific experimental resource for subcellular protein localization (65). Nodes are associated to "Target" through a variety of arrows. GO tems, for example, feature several protein properties, whereas STRING (20) loops protein protein interactions (PPIs) for "Targets"; other resources such as HPA, the Human Protein Atlas (66), informs a variety of nodes regarding tissue, cell type and subcellular location and expression.

**KG-based ML model interpretation.** The AD and ATG XGBoost meta-path models exhibit external predictivity, as some of the top 20 genes predicted by each model have demonstrated valid associations. Some valid associations were also found among the bottom-ranked genes in each model. Since the XGBoost algorithm generates a gradient (probability) output and input labels are binary (1 for positives and 0 for negatives), examining both top- and bottom-ranked genes is a logical approach. Three of the bottom-ranked genes displayed signals for both AD and ATG. It is advisable to consider both top- and bottom-ranked genes because machine learning output is not absolute. The distinction between top-predicted and bottom-predicted genes is more evident in their similarity network. Top genes are more likely to be associated with AD or ATG, while the bottom-ranked genes seem less organized, as indicated by STRING enrichment analyses. Both models used an arbitrary set of over 3,000 negative-label genes; however, no experiments have confirmed these genes as true negatives for either ATG or AD. The absence of a "ground truth" (verified true positives and true negatives) likely contributes to the increased complexity and variance observed in KG-based ML research.

One potential complication in knowledge graph-based machine learning models is "data leakage" (74), which can occur inadvertently due to the complexity of the dataset. In the context of the ATG-based ML model, the inclusion of gene ontology (GO) (64) and Kyoto Encyclopedia of Genes and Genomes (KEGG) (75) terms can introduce information directly related to the research question. Both GO and KEGG contain terms associated with autophagy. Post-ML queries for ATG dark genes led to the exclusion of one bottom-ranked and two top-ranked genes, as they were annotated with ATG-related GO terms. Data leakage has a significant impact, and Kapoor and Narayanan describe it as a reproducibility crisis in ML science (76). In their analysis, data leakage was detected in 15 of the 20 subject areas associated with biology or medicine.

### 3.3. Hit generation and lead optimization

**Bioactivity and ADMET property ML.** QSAR (Quantitative Structure-Activity Relationships) is a fundamental aspect of computer-aided drug design and has been increasingly incorporated into artificial intelligence in drug discovery, including generative chemistry and multi-property optimization (MPO). Developed from the work of Hansch and Fujita in the 1960s (77), QSAR was the first systematic application of machine learning to design small molecules with desired properties (78). QSAR can cover all aspects of "drug property," ranging from target-based bioactivity to ADMET properties, such as solubility, permeability, and other physico-chemical parameters. QSAR methods are algorithm-agnostic and use descriptors (features) derived from molecular structures (objects) (79) to explain the relationship between changes in molecular structure and the molecular property of interest. QSAR has applications in fields like synthesis planning, materials science, nanotechnology, and clinical informatics (80). QSAR technologies are an integral part of the technologies deployed to hunt for anti-COVID-19 therapies (81). QSAR methodologies can be automatically deployed against thousands of target bioactivities (82). Recent lessons shared by scientists at AstraZeneca (83), Bayer (84), Merck (85) and GSK (86) highlight the complexities of ADMET property prediction (87).

Traditionally, QSARs and other small-molecule property ML models use chemical structures as input, often encoded as SMILES (Simplified Molecular Input Line Entry System) (88). A recent development is the concept of Graph Convolutional Neural Networks (GCNNs) introduced in 2015 (89). GCNNs have been used to model drug combinations (90), drug-target interactions (91), and multiple property predictions (92). Rigorous analyses suggest that GCNNs do not improve predictive performance compared to "classical QSAR" fingerprints (92).

**Generative chemistry.** With the foundational role of QSAR in computer-aided drug design and its wide range of applications, the field has continued to evolve and incorporate more advanced techniques, such as generative chemistry (93). The development of ultra-large chemical libraries (94) and generative chemistry methods, particularly those based on Generative Adversarial Networks (GANs), have significantly expanded the possibilities for hit generation and drug discovery. GANs (95), inspired by game theory and deep learning, have found applications in chemistry, such as the first deep learning GAN used in anti-cancer compound design (96). This application employed a 7-layer adversarial autoencoder architecture with growth inhibition percentage as the discriminator. The ML step in this workflow used NCI-60 data (https://bit.ly/nci60), $GI_{50}$ (growth inhibition of 50%), and compound log concentration (LCONC) data for 6252 compounds profiled on the MCF-7 cell line, focusing the GAN on better $GI_{50}$ at lower LCONC. Using 32 probability vectors, 72 million compounds from PubChem (97) were screened, with a maximum of 10 hits per vector. Over one-third of the selected compounds demonstrated activity in cancer assays or were patented for their anti-cancer potential (96). Building upon this method, a GAN was trained to generate *JAK3*-selective molecules using an entangled conditional adversarial autoencoder (98). This led to the identification of a low-micromolar *JAK3* inhibitor, which was considerably less potent on *JAK2*, *BRAF*, and *RAF1*. Additional generative chemistry platforms are described below.

**Multi-property optimization.** Mathematical modeling of multi-objective decision-making has been prevalent since the 1960s (99). At that time, the process of evaluating alternatives based on predefined objectives or criteria was represented as n-dimensional vectors, which were subject to Pareto optimization, aiming to find non-dominant solutions. Hopfinger (100) framed this discussion in terms of computer-aided drug design in 1985. Pareto optimization has emerged as the preferred method for MPO, as it allows for the identification of trade-offs between objectives, leading to the discovery of non-dominant solutions (101). This is particularly useful when evaluating drug properties that are anti-correlated. For instance, increasing drug permeability by one log unit can lead to a decrease in water solubility of up to two log units (102). Therefore, finding non-maximal solutions for each drug property is often necessary when optimizing drug properties.

The use of computers for MPO of drug starting points such as hits and leads, at industrial scale began to gain traction two decades ago (103). The first automated drug discovery implementation was disclosed in 2012 (104), which later became the founding technology of ExScientia, an AI-based drug design company. This particular MPO platform encodes target and off-target bioactivities, ADMET properties, and other criteria. It uses a Laplacian-modified (105) Bayesian classifier (106) trained for polypharmacology profiling and employs a vector scalarization procedure (107) for Pareto-based adaptive optimization. TorchDrug is an open-source AI4DD platform (108) built on PyTorch (109), which leverages the power of deep learning to identify novel drugs. It supports property prediction, pretrained molecular representations (110), generative chemistry, and biomedical knowledge graph reasoning (111). Another platform, REINVENT (112), supports MPO by combining reinforcement learning for the generative chemistry model with diversity filters (113) and a flexible scoring function (114). REINVENT relies on PyTorch and RDKit (115) as a chemistry engine, uses predictive ML models implemented in the scikit-learn library (116), and utilizes Tensorboard's implementation (117) in PyTorch for detailed documentation of chemical space navigation processes.

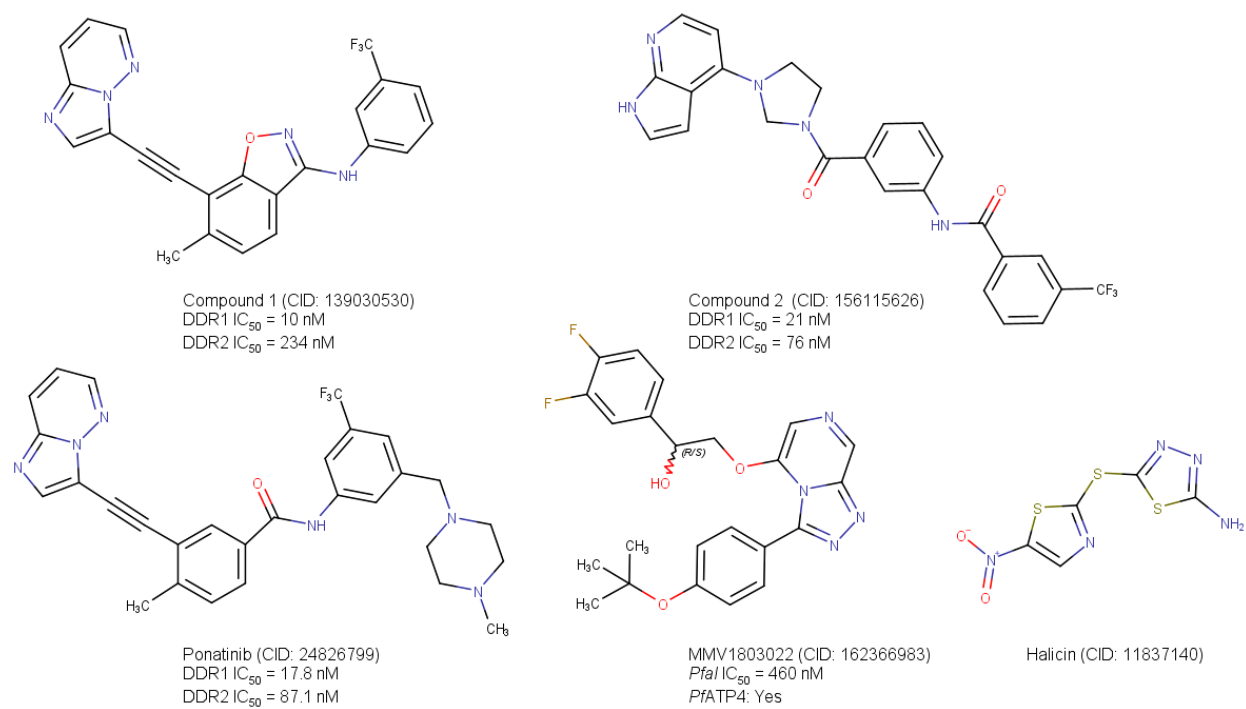

**Figure 2.** Chemical structure of two epithelial discoidin domain-containing receptor inhibitors (compounds 1 and 2), developed using the GENTRL (118) generative chemistry approach. Their chemical similarity with ponatinib (bottom left) is discussed in the text. The structure and bioactivity of MMV1803022, one of the community-driven anti-malarial AI competition compounds, is shown bottom center. *Pfal* summarizes the *in vitro* growth inhibition of asexual blood-stage P. falciparum (3D7) parasite maturation; *Pf*ATP4 indicates confirmed on-target biochemical (cytosolic [Na+] increase) inhibition. Halicin is on the bottom right (see text for details). Bioactivities were extracted from (118) for Compounds 1 and 2, from DrugCentral (47) for ponatinib, and from (119) for the anti-malarial compound. CIDs are PubChem compound identifiers.

GENTRL (Generative Tensorial Reinforcement Learning) is a GAN method that was used to generate *DDR1* and *DDR2* inhibitors (118). DDR (discoidin domain-containing receptor) kinases 1 and 2 are collagen receptors (120). DDR1 plays roles in tumorigenesis, metastasis (121), and fibrosis (122), which makes it an attractive drug target. GENTRL was trained on the "clean leads" subset of the ZINC database (123), enriched with known active and inactive kinase inhibitors from ChEMBL (124). Additionally, it was trained on patented chemical structures, structural (X-ray) data, and DDR1/DDR2-specific pharmacophores. The initial output from GENTRL consisted of 30,000 compounds, which were filtered and prioritized to eliminate undesirable scaffolds, reactive groups, and compounds unlikely to be kinase inhibitors. This process resulted in six novel compounds predicted to be *DDR1* and *DDR2* inhibitors. After synthesis and experimental testing for inhibitory activity, two compounds, Compound 1 and Compound 2, were found to be potent (nM range) inhibitors of DDR1 and DDR2, respectively (Fig 2). The entire process, from nominating *DDR1* to generating Compound 6, took only 35 days. Further experiments demonstrated that Compound 1 is orally bioavailable in mice, with a half-life of 3.5 hours after a 15 mg/kg dose (118).

**Community engagement.** Therapeutic Data Commons (TDC, https://tdcommons.ai/) hosts multiple ML-ready datasets to advance AI4DD technologies systematically (125). TDC features leaderboards for 22 ADMET properties, 5 drug combination benchmarks, one docking, and one drug-target-interaction dataset (29 in total), building on earlier benchmarks for molecular machine learning (126). For example, the MolE deep learning foundational model (127) demonstrates superior performance in 9 of the 22 ADMET property datasets. The QSAR community widely uses OCHEM (ochem.eu), an online platform for ML model development and storage (128). OCHEM provides a uniform interface for multi-task learning (129), a method that promises improved prediction performance (130). Some target-based multi-property ML models are available on-line: NCATS Predictor features 1180 different models (https://predictor.ncats.io/) (131), while REDIAL-2020 (http://drugcentral.org/Redial) estimates twelve properties to evaluate compounds active against COVID (132). For ADMET, one could consider the ADME@NCATS (https://opendata.ncats.nih.gov/adme/) portal (133) or ADMETlab 2.0 (https://admetmesh.scbdd.com/), which includes not only ADMET properties, but also parameters that may aid medicinal chemists (134). Multiple academic and industrial teams participated in an open AI4DD anti-malarial drug competition (119) by targeting *Pf*ATP4, the essential P-type ATP-ase from *Plasmodium falciparum.* MMV1803022, a PfATP4 inhibitor, causes cytosolic alkalinization (via [Na+] increase) and is active against *P. falciparum* (119). See Fig 2.

The output of generative chemistry necessitates human supervision. Following the generative step, active filters are needed to remove potentially reactive and chemically unfeasible species and compounds that fall outside the desired property space (135). Even AI-designed compounds that have been successfully tested can face criticism due to their lack of novelty. For example,

Compound 1 (Fig 2) has been criticized for its similarity to the approved kinase inhibitor, ponatinib (136). An AI-designed antibiotic, "halicin" (Fig 2), disclosed in 2020 (137), has twelve "active" reports in PubChem BioAssays (https://bit.ly/halicinAssays), including an "active" antibiotic result against *Mycobacterium tuberculosis* from 2017. According to *Chemical & Engineering News*, researchers "chose not to pursue [halicin] because of its similarity to a compound that the US Food and Drug Administration was already evaluating" (138). These cases highlight the need for improved AI-generated compound novelty in drug discovery. These examples underscore the importance of enhancing the novelty of AI-generated compounds in drug discovery.

### 3.4. Drug Safety: The road to clinical trials

**Predictive toxicology.** Small molecule MPO design (discussed earlier) includes toxicity prediction of *in vitro* outputs related to organ toxicities such as cytotoxicity and mitochondrial toxicity, or representative assays for predicting off-target toxicities, as well as genotoxicity (139). New advancements include analyses of novel data types such as gene expression data and data from cell imaging experiments, in combination with chemical structure information, to predict *in vivo* toxicity-related effects (140). Using biological fingerprints, either alone or in combination with structural fingerprints, can better relate compounds to *in vivo* phenotypes, sometimes with superior predictivity (141). Including a biological matrix has the advantage of better identifying the subtle differences in phenotype not fully captured by structural similarity alone (142) and can be further facilitated by the inclusion of imputed values in the biological matrix. An example is the application of a neural network applied to fill a sparse kinase binding matrix (143). An important aspect of predictive toxicology is toxicity in context of exposure and understanding the safety threshold. Bayesian ML approaches have proved useful for predicting clinical outcomes from mechanistically relevant *in vitro* data combined with animal exposure (144). Furthermore, AI-augmented Clinical Pharmacology approaches can impact dose recommendation, drug-drug interaction prediction, and adverse drug reaction prediction (145) by utilizing digital data sources such as electronic health records, biomarker data, and patient genomics data. KG-based MLs that incorporate side effect information from package inserts, target information, and indication information have shown promise in predicting adverse events (146).

**Regulatory applications.** Regulatory agencies are showing increasing interest in AI4DD and its uses. The application of computational tools to evaluate the carcinogenic potential of drug-related impurities has led to the inclusion of structure-activity predictions as part of regulatory submissions (147). In 2018, the European Medicines Agency published a reflection paper discussing the use of *in silico* tools for assessing risks related to non-mutagenic impurities when experimental data is unavailable (148). The use of AI in regulatory applications introduces a new aspect, requiring models to be transparent enough for Health Authorities to assess their usability and reliability. Cross-industry consortia have been working towards formalizing the use of *in silico* tools, including AI/ML models, to ensure transparency in data set origins and quality, as well as algorithms, in order to gain regulatory acceptance (149, 150).

AI is gaining traction in various applications, such as chemical risk assessment related to occupational toxicology, transportation of chemicals, and medical devices, where endpoints like skin sensitization, irritation, and rat acute oral toxicity are commonly included (151, 152). In the area of Modeling and Simulation (M&S), which is now a standard tool used to demonstrate effects on physiology or safety when expanding to other clinical populations or indications, efforts are being made to define criteria for model evaluation (153). The National Center for Toxicological Research (NCTR), in collaboration with the Center for Drug Evaluation and Research (CDER), has launched the "SafetAI" initiative to develop AI models for toxicological endpoints crucial for assessing drug safety, potentially playing a role in the IND review process (154). This project focuses on developing deep learning-based models for hepatotoxicity, carcinogenicity, nephrotoxicity, among others. It utilizes various *in vitro* data sources, such as high-throughput transcriptomics, in addition to chemical representation to predict *in vivo* outcomes.

## 4. Brief overview of AI-driven drug discovery successes

The first AI-designed compound is DSP-1181, a potent and long-acting 5HT$_{1A}$ receptor agonist (https://www.exscientia.ai/dsp-1181). DSP-1181 was designed using ExScientia's MPO approach, Centaur Chemist, in less than 12 months. Sumitomo Dainippon Pharma progressed DSP-1181 into phase I clinical studies for obsessive-compulsive disorder in January 2020. The chemical structure of DSP-1181 has not been disclosed (155). Chemical Abstracts Service analysts suggested DSP-1181 is similar to haloperidol (https://bit.ly/3KePr9u). Since haloperidol is a weak 5HT$_{1A}$ antagonist (data not shown), we suggest it is similar to gepirone, a low-efficacy 5HT$_{1A}$ agonist. Another AI-designed compound, DSP-0039, is a dual 5-HT$_{1A}$ agonist and 5-HT$_{2A}$ antagonist with no dopamine D2 receptor activity (156). Sumitomo Dainippon Pharma progressed to phase I clinicals for treatment of Alzheimer's disease psychosis in May 2021. See Fig 3 for details.

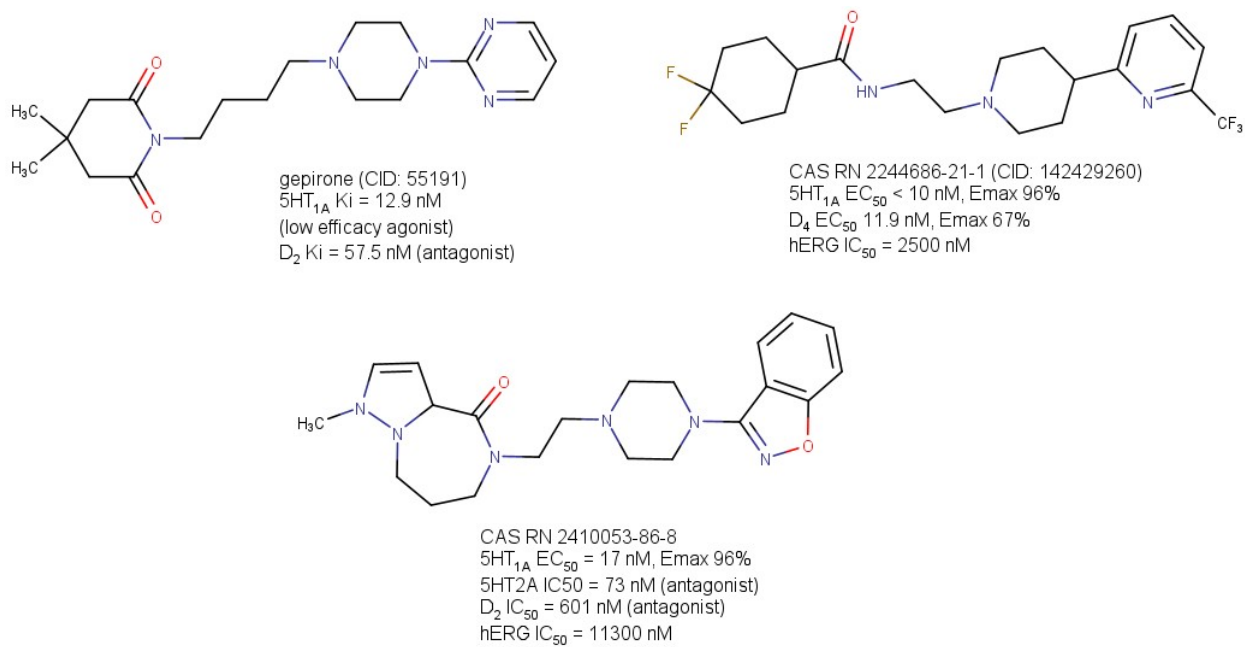

**Figure 3.** Examples of compounds reported in ExScientia patents. Gepirone (top left) shares the pharmacological profile of Example 1 (CAS RN 2244686-21-1; top right). Example 109 (CAS RN 2410053-86-8; bottom) does not share chemical similarity with known drugs. Bioactivity data for patented examples is from (155), (156) and DrugCentral for gepirone (47). CIDs are PubChem compound identifiers.

ExScientia has made significant AI-driven advancements in functional precision oncology. A single-cell functional precision medicine (scFPM) platform successfully guided treatment selection and enhanced patient outcomes in a prospective clinical study (157). Utilizing the scFPM approach, 139 drugs were tested on samples from 143 patients with hematologic malignancies. Out of the 56 patients treated based on scFPM results, 30 (54%) experienced over 1.3-fold increased progression-free survival compared to their previous therapy. Additionally, 12 (40% of responders) had exceptional responses lasting three times longer than expected for their respective diseases (157). Compared to patients receiving physician-chosen therapy, those treated with scFPM demonstrated a notable overall survival benefit.

Insilico Medicine's ISM01-055 (also known as ISM018-055) may be the first AI-designed compound to enter phase II clinical trials in June 2023 (https://bit.ly/46BpZ7m). Designed using the PandaOmics (158) and Chemistry42 (159) platforms, ISM018-055, which may be covered by this patent (160), targets NCK-interacting protein kinase, *TNIK* (http://bit.ly/insmhkexa1) and is indicated for the treatment of idiopathic pulmonary fibrosis.

Relay Therapeutics, an AI4DD company focused on molecular dynamics, disclosed RLY-1971 (Fig 4), an orally bioavailable allosteric *PTPN11* (SHP2) inhibitor. RLY-1971 blocks wild-type *PTPN11* ($IC_{50}$ <1nM) and the E76K activating mutant ($IC_{50}$ <250nM) and is in phase I clinical trials for the treatment of RTK/RAS driven solid tumors (161). RLY-1971 was licensed by Genentech in December 2020 (https://bit.ly/3NWzarS). RLY-4008, a highly selective, potent and irreversible FGFR2 inhibitor, is effective in cholangiocarcinoma (162). See also Fig 4.

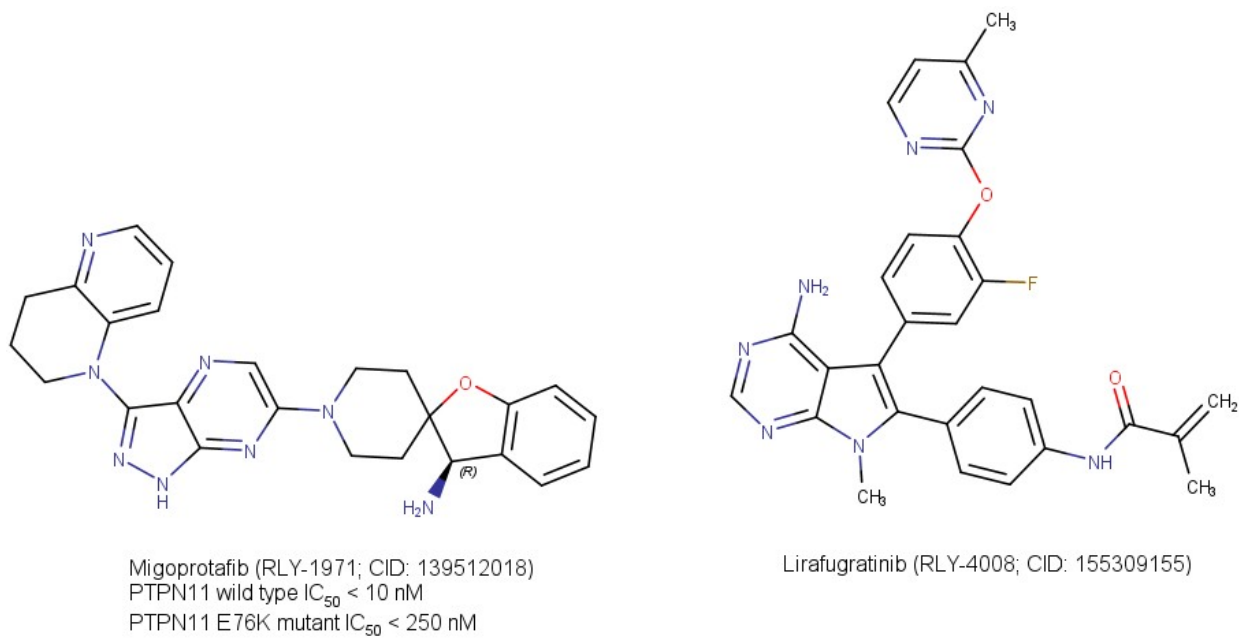

Figure 4. Chemical structures of RLY-1971 (migoprotafib or GDC-1971) and RLY-4008 (lirafugratinib), two compounds progressed into phase I clinical studies by Relay Therapeutics. While the exact bioactivity of RLY-4008 has not been disclosed, it lacks potency against FGFR1 and FGFR4 (163), which in turn may eliminate unwanted side-effects. CIDs are PubChem compound identifiers.

## 5. Challenges and limitations of AI in drug discovery

"Artificial intelligence in drug discovery" might give the impression that AI is successfully employed in early drug discovery. While there is some truth to this statement, it's important to note that no medicines approved by regulatory agencies can be attributed to AI in the same way AI has achieved victories in chess, Go, *Jeopardy!* autonomous vehicles, or poetry generation. Drug discovery is a complex, multifaceted process, as captured by the 4DM charts. Robot Scientist Adam independently conducted genomic experiments (164), and Robot Scientist Eve performed an HTS campaign to identify anti-malarial compounds (165). Although computer-aided processes have been used for compound selection and optimization, no AI-driven Robot Scientist or digital equivalent currently exists that can execute fully automated drug discovery. Automated AI-driven drug discovery remains an aspirational goal (166). Most success stories to date have relied on machine learning, cheminformatics, bioinformatics software, natural language processing, or other computational platforms that support human decision-making. In summary, drug discovery has yet to benefit from a comprehensive AI system.

One of the weak aspects of AI4DD for small molecules is the training of ML models that encode chemical features (often referred to as QSARs), such as those derived from chemical structures. Bohacek et al. estimated (167) the number of 'drug-like' chemicals to be up to $10^{60}$, and virtual screening libraries have already exceeded 30 billion compounds (94). The logistical and practical challenges of virtually screening 30 billion compounds, considering an estimated 10-50 conformers per molecule, amounting to nearly 500 billion objects, are immense and beyond the scope of this review. Instead, our focus is on the practical issues related to the applicability domain (168) and external predictivity validation (169). Both validation and applicability are challenges faced by target-based KG machine learning models, as mentioned earlier.

Machine learning models commonly used in AI4DD are often trained on tens of thousands of compounds or less, which raises questions about their effectiveness in sampling the chemical space of 30 billion compounds. Can we confidently assume that such comparatively small training sets effectively represent the chemical space of 30 billion? Are such predictions trustworthy? Both the applicability domain (a representational issue in ML feature space) and chemical diversity (unseen scaffolds are less likely to produce reliable predictions even within the applicability domain) raise concerns about the predictivity of ML models for the unexplored "chemical universe." Ideally, AI4DD practitioners would want to systematically sample chemical space using adequately trained ML models. This becomes imperative during lead optimization, where progress depends on the accurate representation of relevant chemical scaffolds in the ML space.

Target identification in drug discovery is also impacted by the so-called reproducibility crisis (170). Bayer (171) and Amgen (172) have reported low reproducibility rates (33% and 11%, respectively) of high-impact publications, and many biomedical publications are false (173). From an AI perspective, filtering out false data requires a coordinated community effort. Lessons (174) from *eLife's* Reproducibility Project, which focused on cancer biology, highlight issues like weaker-than-previously-published (175) effects and inaccurate protocol descriptions (176), among others. The possibility of indexing fabricated publications (177) or those generated by "research paper mills" (178) further increases the likelihood of false information in the field. These challenges with experimental data compound the issues of ML model accuracy and the ML science reproducibility crisis (76). For AI4DD to be effective, it needs to be anchored in truth.

Another subtle risk involves the education of scientists in using AI4DD models effectively. Questions about when, how, and in what order to deploy ML models are crucial. The proper use of ML models depends on the specific requirements of each unique drug discovery project. For some projects, target selectivity and appropriate in-tissue delivery might be more important than absolute affinity or systemic toxicity. In contrast, other projects might focus on mitigating on-target toxicity, low permeability, or scaffold similarity to competitor patents. Each issue demands different computational solutions, ranging from filters and lead hopping to sequential ML model deployment. Proper training in using AI4DD models is critical to ensure that scientists can effectively navigate these complexities and make informed decisions based on the specific needs of their drug discovery projects.

The human component of drug discovery is another crucial aspect. In many academic and industrial settings, decision-making falls to medicinal chemists who typically rely on their judgment to propose compounds for the design-make-test (DMT) cycle rather than depending solely on AI. Compounds may be thoroughly evaluated by the project team, with members voting on the order in which chemicals should be synthesized and tested. In AI-integrated companies, AI may influence this process, but chemists are still likely to veto compounds that don't meet specific criteria, even if computational chemists or toxicologists find no issues. It is reasonable to assume that user expertise, bias, and time constraints play a significant role in early drug discovery, often more so than AI. The Pfizer "rule of 5" (Ro5 or Lipinski rules) serves as an early example (179) of attempts to integrate informatics and data science into the early stages of drug discovery. The Ro5 criteria, assessing hydrogen bonding capacity, the calculated octanol/water partitioning coefficient (logP), and molecular weight have been employed world-wide to narrow down the chemical solution space. It is undeniable that Ro5 criteria have had a significant impact on medicinal chemistry (180). However, the influence of these criteria is gradually diminishing over time (181).

## 6. Conclusions and future outlook

Given the knowledge deficit caused by the constantly-increasing volume of information, AI systems capable of independent reasoning are becoming a necessity. AI has making significant advances in disease diagnosis and healthcare professions, such as radiology, pathology, and clinical pharmacology. Although nosology remains a challenge for both humans and AI systems, resources like the Monarch Disease Ontology demonstrate the potential for human and computer collaboration. In target identification, KG-based machine learning can structure, process, and evaluate large scale datasets, with many potential applications in biomedicine. However, KG-based ML models face challenges like data leakage and the absence of ground truth, which may contribute to increased variance in research. Integrating generative chemistry, ML and MPO techniques like Pareto optimization are emerging as integral components of AI methods for hit and lead discovery, with strong support from community-based platforms that provide datasets and resources for AI4DD technologies. Significant advances are taking place in predictive toxicology, and regulatory agencies are positioning themselves to benefit from AI technologies. We also discussed several AI-designed compounds that have entered phase I clinical trials. These success stories rely on computational platforms to support human decision-making, and AI has yet to deliver a comprehensive system for fully automated drug discovery. Machine learning models used in AI4DD face challenges related to applicability domain, external predictivity validation, and chemical diversity. Proper training in using AI4DD models is crucial to ensure effective navigation of complexities and informed decision-making in drug discovery projects, with human expertise playing a significant role.

The reproducibility crisis - both for experiments and ML models - suggests that the scientific community is responsible for laying down the foundations of verity and carefully vetting what is known. Modern science has no "we hold these truths to be self-evident" basis. Biology does not appear to lend itself to an axiomatically built foundation. But managing AI4DD systems implies knowledge and model update, and we collectively share the responsibility of sifting truthful science from questionable results. The true strength of any AI4DD system lies in its ability to process and comprehend sparse data more efficiently and effectively than humans. Computers are better suited to process and interpret complex molecular and knowledge graph representations compared to humans. However, this strength can only be revealed when "true data" is provided. The adage, "garbage in, garbage out," is a stark reminder of the need for "ground truth" across all data types and relationships in drug discovery.

One of the potential dangers is the misguided use of AI4DD tools. The prevailing notion is that feeding ML models with massive data can produce actionable results. This approach, destined to fail, needs rethinking. A better scenario is to allow AI to steer the entire drug discovery process, particularly in structure-based projects or when there is a wealth of bioactivity data. Human intervention could occur only at the very late stages of the pipeline. By fully leveraging the potential of AI, the drug discovery process can be significantly optimized and revolutionized.

## 7. Epilogue

ChatGPT, the conversational AI that has passed the US medical licensing board, has the potential "to revolutionize research practices and publishing" (182). The GPT-4 technical report from March 14, 2023, demonstrates how GPT-4 could be used to create new drugs, among other applications (183). In one use-case scenario, Andrew White, a member of OpenAI's "red team," prompted GPT-4 with the name Dasatinib, a kinase inhibitor drug. GPT-4 was asked to modify the drug and find novel, non-patented molecules with a similar mode of action, locate chemical vendors selling the compound, and purchase it. If custom synthesis was needed, GPT-4 was to email a contract research organization to order the compound. The prompt, instructions and compound chosen by GPT-4 can be found on page 59 of the OpenAI manuscript.

Notable observations from the experiment include: 1) GPT-4 generated a valid chemical structure (SMILES (88)) output, indicating its ability to perceive and modify chemical structures correctly; 2) the molecule is available in the ZINC database (123), meaning it is synthetically feasible; 3) the proposed molecule is desmethyl-imatinib, an N-dealkylated piperidine metabolite of imatinib, another protein kinase inhibitor drug. GPT-4 successfully modified the molecule while retaining its kinase inhibitor properties. Experimental validation is needed to confirm if the GPT-4-generated molecule shares the same mode of action as Dasatinib. It is essential to remember that GPT-4 has general expertise and is not specifically tailored for drug discovery. By providing access to external chemistry tools, White et al. have designed ChemCrow, a GPT-4 based tool that demonstrates how appropriate technologies can significantly enhance LLM effectiveness in chemistry-related tasks (184). If more external resources are integrated with GPT-4 or its successors, LLMs could become a valid approach.

In the Dreamworks animated movie "Shrek", Donkey keeps asking: "Are we there yet?" and Shrek and Fiona keep saying "No!". To those waiting for AI4DD: We're not there yet but we are on our way!

**Acknowledgments:** We would like to thank Melissa Landon, Alexander Tropsha, John Overington, Miguel Costa Coelho, Christopher Southan and Andrew White for discussions about this topic. Jessica Binder helped design Figure 1.

**Disclosure:** Catrin Hasselgren is a full time employee of Genentech Inc. Tudor Oprea is a full time employee of Expert Systems Inc. Tudor is a member of the Scientific Advisory Board for In Silico Medicine, one of the companies featured in this paper. Part of this work was supported by NIH Grant NCI U24 224370.